\documentclass[lettersize,journal]{IEEEtran}
\usepackage{amsmath,amsfonts}
\usepackage{array}
\usepackage{textcomp}
\usepackage{subfigure}
\usepackage{url}
\usepackage{verbatim}
\usepackage{graphicx}
\usepackage{multirow}
\usepackage{booktabs}
\usepackage{float}
\usepackage{cite}
\usepackage{color}
\usepackage{textcase}
\usepackage{color}
\usepackage{listings}
\usepackage{threeparttable}
\def\BibTeX{{\rm B\kern-.05em{\sc i\kern-.025em b}\kern-.08em
    T\kern-.1667em\lower.7ex\hbox{E}\kern-.125emX}}
\usepackage{balance}
\usepackage[colorlinks,bookmarks=false,citecolor=green]{hyperref}
\usepackage{algorithm}
\usepackage{algorithmicx}
\usepackage{algpseudocode}

\begin{document}
\title{Cross-Model Cross-Stream Learning for Self-Supervised Human Action Recognition}

\author{
Mengyuan Liu, Hong Liu, Tianyu Guo

\thanks{M. Liu, H. Liu and T. Guo are with the State Key Laboratory of General Artificial Intelligence, Peking University, Shenzhen Graduate School. This work was supported by Natural Science Foundation of Guangdong Province (No. 2024A1515012089), National Natural Science Foundation of China (No. 62203476). Corresponding authors: Hong Liu (hongliu@pku.edu.cn) and Tianyu Guo.}
}

\maketitle
\begin{abstract}
Considering the instance-level discriminative ability, contrastive learning methods, including MoCo and SimCLR, have been adapted from the original image representation learning task to solve the self-supervised skeleton-based action recognition task. These methods usually use multiple data streams (i.e., joint, motion, and bone) for ensemble learning, meanwhile, how to construct a discriminative feature space within a single stream and effectively aggregate the information from multiple streams remains an open problem. To this end, {this paper} first applies a new contrastive learning method called BYOL to learn from skeleton data, {and then} formulate SkeletonBYOL as a simple yet effective baseline for self-supervised skeleton-based action recognition. Inspired by SkeletonBYOL, {this paper} further presents a Cross-Model and Cross-Stream (CMCS) framework. {This framework} combines \textcolor{black}{C}ross-\textcolor{black}{M}odel \textcolor{black}{A}dversarial \textcolor{black}{L}earning (CMAL) and \textcolor{black}{C}ross-\textcolor{black}{S}tream \textcolor{black}{C}ollaborative \textcolor{black}{L}earning (CSCL). Specifically, CMAL learns single-stream representation by cross-model adversarial loss to obtain more discriminative features. To aggregate and interact with multi-stream information, CSCL is designed by generating similarity pseudo label of ensemble learning as supervision and guiding feature generation for individual streams. {Extensive} experiments on three datasets verify the complementary properties between CMAL and CSCL and also verify that {the proposed method} can achieve better results than state-of-the-art methods using various evaluation protocols. \end{abstract}


\begin{IEEEkeywords}
Skeleton-Based Action Recognition, Self-Supervised Learning, Multi-Stream.
\end{IEEEkeywords}

\section{Introduction}
\IEEEPARstart{H}UMAN action recognition has wide applications in human-computer interaction, video surveillance, and health care \cite{tu2019action, liu2022generalized, zhang2022zoom, hadikhani2023novel, taghanaki2023self, wang2022predicting, baruah2023intent, tu2022joint,tu2023consistent,gao2023dual}. Different from RGB videos or depth data, 3D skeleton data encodes high-level representations of human actions and it is robust to background clutter, illumination changes, and appearance variation \cite{tu2022joint, liu2023novel, liu2023temporal}. In addition, with the popularity of depth sensors \cite{kinect} and the development of advanced human pose estimation algorithms \cite{openpose,alphapose,simplebaseline,videopose,strided}, the acquisition of skeleton data has become accessible. Thus, skeleton-based action recognition has gradually attract more and more attention.

Previous works have explored various approaches for skeleton-based action recognition. Traditional methods \cite{Actionlet, Lie-Group, Rolling} focus on hand-crafted features to model skeleton sequence patterns. With the advancements in deep learning, Recurrent Neural Networks (RNNs) and Long-Short Term Memory (LSTM) have been used to model temporal dynamics \cite{HB-RNN, ST-LSTM, VA-RNN}. Further, Convolutional Neural Networks (CNNs) based methods \cite{XYZ1, XYZ2, cao2018skeleton, banerjee2020fuzzy} convert skeleton sequences to RGB images and use CNNs to learn spatio-temporal features. In recent years, due to the similarity between the human skeleton and graph structure, some methods \cite{ST-GCN, 2s-AGCN, MST-GCN} capture spatiotemporal relationships leveraging the graph structure of human skeletons based on Graph Convolutional Networks (GCNs), which have achieved superior performances. Nevertheless, above methods are all trained in a supervised manner, which is costly to obtain for large-scale datasets. To address this issue, self-supervised skeleton-based action recognition has gained attention \cite{LongGAN,PandC,AS-CAL,crossclr,isc,MS2L,CP-STN,WACV,PCRP} as it can better utilize unlabeled data to learn more generalized feature representations.

In the domain of self-supervised skeleton-based action recognition, certain approaches \cite{LongGAN,PandC} employ generative pretext tasks, such as reconstruction and prediction, to learn representations from unlabeled skeleton sequences. On the other hand, as contrastive self-supervised learning gains popularity, other works \cite{AS-CAL,crossclr,isc} adopt the contrastive learning framework to achieve better results in this field. They apply data augmentations on the skeleton sequence to construct contrastive pairs (i.e., positive pairs and negative pairs) and hope to learn general representations by pulling the positive pairs closer and pushing the negative pairs away. There are also some methods \cite{MS2L, CP-STN, WACV, PCRP} that combine the generative task with the contrastive task to extract discriminative representations in the multi-task learning manner.

Among them, compared with the generative methods, the contrastive methods pay more attention to the instance-level information rather than the detailed information and then construct a more discriminative feature space that is more suitable for downstream tasks, so it has received extensive attention. Presently, the contrastive methods utilized in self-supervised skeleton-based action recognition \cite{crossclr,AS-CAL,isc} require special attention when handling negative samples. This is achieved through techniques like using large batch sizes, memory banks, or negative sampling strategies to effectively retrieve and utilize negative samples during training. These approaches help enhance the learning process and improve the quality of learned representations for the task. {Different from previous methods, BYOL \cite{byol} does not use negative samples, and learns better feature representation through a simple framework, which provides a new idea for the development of contrastive learning in the field of skeleton action recognition.
Therefore, this paper adopts BYOL as the baseline to build our method.}
{Further, this paper rethinks existing contrastive self-supervised skeleton-based action recognition methods and find several unsolved problems.}

\begin{figure}[t]
\centerline{\includegraphics[width=8.8cm]{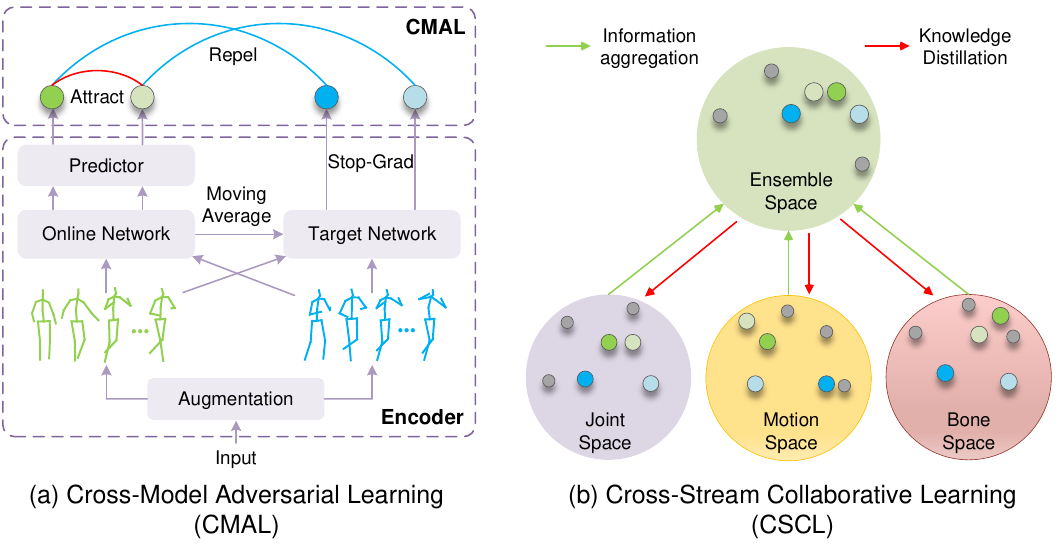}}
\vspace{-1em}
\caption{Illustration of our Cross-Model Cross-Stream (CMCS) framework, which mainly involves an Encoder block, a Cross-Model Adversarial Learning (CMAL) block, and a Cross-Stream Collaborative Learning (CSCL) block. (a) For a single stream, better intra-stream representation is learned by CMAL with ``Attract'' and ``Repel'' operations on features extracted from an Encoder. {To facilitate observation, two skeleton sequences are colored in green and blue. The green dot means the feature extracted from the green skeleton sequence and the blue dot means the feature extracted from the blue skeleton sequence.} (b) For multiple streams, CSAL aggregates inter-stream information and makes the feature space of a single stream to be consistent with the ensemble space. (Best viewed in color)}
\label{motivation}
\vspace{-1em}
\end{figure}

\textbf{1) How to construct a discriminative feature space within a single stream?} Although the above methods have improved the skeleton representation ability to a certain extent, it is believed that the ability of self-supervised methods has not been fully explored. Using the BYOL framework \cite{byol} directly is not necessarily the most suitable for skeleton action recognition tasks, and the learned feature discrimination could be further improved. Constructing a more discriminative feature space within a single stream is a very challenging problem, especially for the contrastive learning framework without negative samples.

\textbf{2) How to interact and aggregate the information from multiple streams?} In skeleton action recognition, joint, motion, and bone data are three common types of data used to describe human motion. Joint data provides information about the position of each joint in the human body. Motion data describes how the joints are moving over time while bone data describes the spatial relationship between adjacent joints. They are complementary because they each provide unique information and by combining these three types of data, this paper can get a more complete understanding of the action. However, existing methods often use simple ensemble learning, which cannot perform information interaction well. If such complementary information could be fully utilized and explored, a better feature representation would be obtained. Thus, how to better fuse information between streams needs to be carefully considered.

To this end, this paper presents a Cross-Model Cross-Stream (CMCS) framework to enhance both single-stream and multi-stream representations for self-supervised skeleton-based action recognition. Specifically, SkeletonBYOL is first proposed following BYOL \cite{byol} as the baseline. Secondly, Cross-Model Adversarial Learning (CMAL) is proposed to effectively learn single-stream representation as shown in Fig. \ref{motivation} (a). Unlike BYOL, CMAL is proposed to make the feature ``Attract'' or ``Repel'', and then get a more discriminative feature space. Thirdly, Cross-Stream Collaborative Learning (CSCL) is proposed to aggregate cross-stream information as shown in Fig. \ref{motivation} (b). The feature distribution of the three streams (i.e., joint, motion, and bone) is fused into an ensemble similarity pseudo-label and the pseudo-label is used as a supervision signal to optimize the feature space of every single stream. Then the pseudo-label continuously updates and iterates to enhance the representations.

In summary, our main contributions are two-fold:

\begin{itemize}
\item This paper presents a Cross-Model Cross-Stream (CMCS) framework to effectively enhance both single-stream and multi-stream representations for self-supervised skeleton-based action recognition.

\item This paper formulates a simple yet effective baseline, i.e., SkeletonBYOL. Based on the encoder of SkeletonBYOL, our CMCS framework further involves Cross-Model Adversarial Learning (CMAL) and Cross-Stream Collaborative Learning (CSCL) to learn distinctive representations from both single-stream and multi-streams.

\end{itemize}

\section{Related Work} \label{section2}

\textbf{Supervised Methods.} Traditional methods for skeleton-based action recognition \cite{Actionlet, Lie-Group} rely on hand-crafted features. With the advancement of deep learning, some approaches \cite{HB-RNN, ST-LSTM} use RNNs to model the temporal features of skeleton sequences. However, RNNs suffer from gradient vanishing issues. To address this, other methods \cite{XYZ1, XYZ2, PR} convert the 3D skeleton sequence into pseudo images and utilize CNNs, achieving competitive results. Nonetheless, both RNNs and CNNs do not fully capture the inherent graph structure of skeleton data. Recently, GCNs, particularly ST-GCN \cite{ST-GCN}, have gained popularity in skeleton-based action recognition, leading to the emergence of various GCN-based methods \cite{2s-AGCN, MST-GCN, zheng2021cross}. This paper uses ST-GCN as the encoder for feature extraction to demonstrate the effectiveness of our proposed framework.

\textbf{Self-supervised Methods.} Self-supervised learning falls under the category of unsupervised learning and involves defining pretext tasks as learning objectives to acquire meaningful representations from unlabeled data. For the skeleton sequence data, there are generative pretext tasks \cite{LongGAN,PandC,HTR,cloud,MCAE}, contrastive pretext tasks \cite{AS-CAL,crossclr,ST-CL,isc,xu2023spatiotemporal,moliner2022bootstrapped,mao2022cmd,pang2023skeleton}, and multiple pretext tasks. For generative pretext tasks, LongTGAN \cite{LongGAN} employs an encoder-decoder architecture to reconstruct the input sequence and derive meaningful feature representations. P\&C \cite{PandC} also relies on a reconstruction pretext task but adopts a training strategy to weaken the decoder, encouraging the encoder to learn more discriminative features. Cheng et al. \cite{HTR} propose a pretext task involving predicting the motion of 3D skeletons, utilizing the Hierarchical Transformer for encoding the skeleton sequences. Yang et al. \cite{cloud} introduce a novel technique called skeleton cloud colorization to facilitate the learning of skeleton representations. Xu et al. \cite{MCAE} propose the Motion Capsule Autoencoder (MCAE) to address the ``transformation invariance'' challenge in the motion representations. Indeed, methods based on frame-level generation, like those mentioned previously, may not be inherently suitable for representation learning in the context of skeleton-based action recognition \cite{ST-CL}. There might be limitations in effectively capturing the underlying spatiotemporal relationships and structures present in the skeleton data using such frame-level generation approaches.

For contrastive pretext tasks, AS-CAL \cite{AS-CAL} proposes to use plenty of spatiotemporal augmentations and use momentum LSTM and memory bank. SkeletonCLR \cite{crossclr} is built based on MoCo v2 \cite{moco, mocov2} and uses \textit{Shear} and \textit{Crop} as the augmentations, which achieves competitive performance. Further, CrosSCLR \cite{crossclr} introduces a cross-view consistent knowledge mining strategy to enhance the performance of SkeletonCLR. ST-CL \cite{ST-CL} leverages the spatiotemporal continuity of motion tendency as the supervisory signal to acquire action-specific features. Thoker et al. \cite{isc} propose inter-skeleton contrastive learning, which learns from multiple different input skeleton representations in a cross-contrastive manner. Xu et al. \cite{xu2023spatiotemporal} propose the spatiotemporal decouple-and-squeeze contrastive learning (SDS-CL) framework, which jointly contrasts spatial-squeezing features, temporal-squeezing features, and global features while incorporating a new attention mechanism to improve feature learning. Moliner et al. \cite{moliner2022bootstrapped} propose to use vanilla BYOL and asymmetric augmentation, but its gain mainly comes from using a larger number of encoder channels. CMD \cite{mao2022cmd} also uses the MoCo v2 framework and proposes cross-modal mutual distillation to improve the performance. Pang et al. \cite{pang2023skeleton} propose a novel contrastive GCN-Transformer network (ConGT), focusing on designing a stronger encoder.

The generative pretext tasks pay more attention to the detailed information of the skeleton sequence, and the contrastive pretext task pays more attention to the discriminative information of the skeleton sequence at the instance level. In order to learn more general features, there are also several methods to do multiple pretext tasks. MS$^2$L \cite{MS2L} proposes to use multiple pretext tasks (i.e., motion prediction, jigsaw puzzle recognition, and contrastive learning) and combine them to encourage the Bi-GRU encoder to capture more suitable features. CP-STN \cite{CP-STN} integrates contrastive learning paradigms and generative pretext tasks within a single framework by employing asymmetric spatial and temporal augmentations, allowing the network to extract discriminative representations. Tanfous et al. \cite{WACV} propose a taxonomy of self-supervised learning for action recognition, reconciling AS-CAL and P\&C. PCRP \cite{PCRP} introduces a novel framework that employs prototypical contrast and reverse prediction to fully learn inherent semantic similarity in the context of self-supervised action recognition.

Our method belongs to the category of using contrastive pretext tasks. Different from the existing methods\cite{crossclr, AS-CAL, ST-CL}, our method focuses on exploring a new baseline to obtain more discriminative features within a single stream and focuses on how to interact and aggregate information between different streams to obtain a more discriminative feature space than ensemble learning. The proposed method has a simpler structure and is applicable to a variety of encoders.

\section{Method} \label{section3}

\subsection{SkeletonBYOL}

Contrastive learning has gained widespread popularity due to its ability to perform instance discrimination effectively. Motivated by this, we introduce SkeletonBYOL, a method to learn single-stream 3D action representations. Our approach is inspired by {a simple yet effective technique} called Bootstrap Your Own Latent (BYOL) \cite{byol}. SkeletonBYOL is a non-negative samples method for skeleton representation, where each sample is considered as its own positive sample, and we leverage different augmentations of the same sample during training. In each training step, the positive samples are embedded close to each other. SkeletonBYOL consists of four major components.

\textbf{Augmentation.} Data augmentations $\mathcal{T}$ are used to randomly transform the given skeleton sequence $s$ into different augments $x$ and $x'$ that are considered as positive pairs. For skeleton data, \textit{Shear} and \textit{Crop} are adopted as the augmentation strategy following SkeletonCLR \cite{crossclr}.

(i) \textit{Shear}: The shear augmentation is a linear transformation applied to the spatial dimensions of 3D coordinates of body joints. It slants the shape of the joints with a random angle. The transformation matrix for shear is defined as:
\begin{equation}
A = \left[
\begin{array}{ccc}
1      & a_{12} &a_{13}\\
a_{21} &    1   &a_{23}\\
a_{31} & a_{32} &1
\end{array}
\right].
\end{equation}
The shear factors, denoted as $a_{12}$, $a_{13}$, $a_{21}$, $a_{23}$, $a_{31}$, and $a_{32}$, are randomly sampled from a uniform distribution within the range $[-\beta, \beta]$, where $\beta$ is the shear amplitude. The transformation matrix $A$ is then formed with these shear factors and applied to the skeleton sequence along the channel dimension.

(ii) \textit{Crop}: For the crop augmentation, it is commonly used in image classification tasks as it enhances diversity while preserving the distinguishing features of the original samples. {In the case of temporal skeleton sequences, the first several frames are flipped and concatenated to the original sequence. Also the last several frames are flipped and further concatenated to the above sequence. Finally, the sequence is cropped back to the original length.} {Let $T$ denote the length of the skeleton sequence.} The length of padding is defined as $T/\gamma$, where $\gamma$ is the padding ratio, and it is set to 6.

\textbf{Online Network.} The online network is defined by a set of weights $\theta$ and is comprised of three components: an encoder $f_\theta$, a projector $g_\theta$, and a predictor $q_\theta$. The encoder $f_\theta$ embeds $x$ into feature space: $y_\theta = {f_\theta }(x)$. The projector $g_\theta$ is a multi-layer perceptron (MLP), which consists of a linear layer followed by batch normalization, a rectified linear unit (ReLU), and a final linear layer. $g_\theta$ is used to project the feature vector to a new feature space: $z_\theta = {g_\theta }({y_\theta })$. The predictor $q_\theta$ is used to predict the output of the target network, which has the same structure as $g_\theta$. The parameters $\theta$ of the online network are updated via gradients.

\textbf{Target Network.} The target network is comprised of two components: an encoder $f_\xi$ and a projector $g_\xi$. They have the same architecture as that in online network but use a different set of weights $\xi$. The target network provides the regression targets to train the online network: $y'_\xi = {f_\xi }(x')$, $z'_\xi = {g_\xi }({y'_\xi })$. Its parameters $\xi$ are an exponential moving average of the online parameters $\theta$. More precisely, given a target decay rate $\tau \in \left[ {0,1} \right]$, after each training step we perform the update:
\begin{equation}
{\xi} \leftarrow \tau{\xi} + (1 - \tau){\theta}.
\label{eq1}
\end{equation}

\begin{figure}[t]
\centerline{\includegraphics[width=9cm]{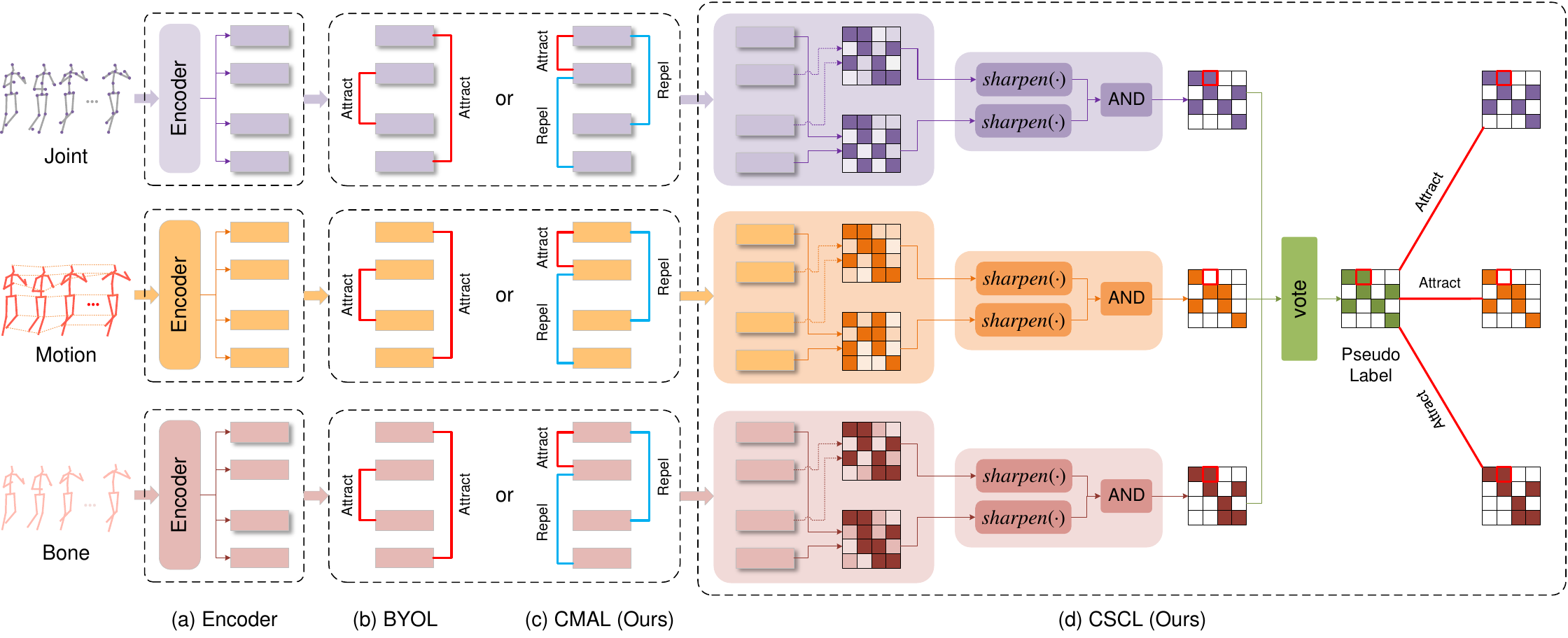}}
\caption{Comparison between our proposed Cross-Model Cross-Stream (CMCS) framework and our proposed simple baseline, i.e., SkeletonBYOL. Our CMCS framework consists of (a) Encoder, (c) CMAL, and (d) CSCL. Meanwhile, the SkeletonBYOL consists of (a) Encoder and (b) BYOL. Noting that the Encoder contains Augmentation, Online Network, and Target Network. {The red line means ``Attract" and the blue line means ``Repel".} (Best viewed in color)}
\label{3s-SkeletonBIISD}
\end{figure}

\begin{figure}[t]
\centerline{\includegraphics[width=9cm]{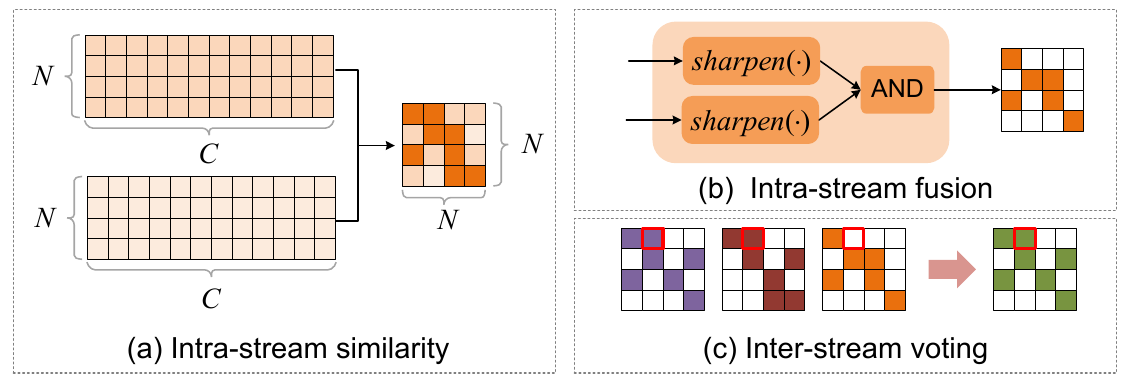}}
\caption{Three key components of our proposed CSCL. (Best viewed in color)}
\label{module}
\end{figure}

\textbf{BYOL.} Finally we define the following mean squared error between the $l_2$-normalized prediction ${q_\theta }({z_\theta })$ and target projections ${{z'}_\xi }$:
\begin{equation}
\begin{aligned}
\mathcal{L}_{\theta, \xi} &= \left\|{{q_\theta }({z_\theta }) - sg({{z'}_\xi })} \right\|_2^2 \\
&= 2 - 2 \cdot \frac{{\left\langle {{q_\theta }({z_\theta }), sg({{z'}_\xi })} \right\rangle }}{{{{\left\| {{q_\theta }({z_\theta })} \right\|}_2} \cdot {{\left\| {sg({{z'}_\xi })} \right\|}_2}}},
\end{aligned}
\label{eq2}
\end{equation}
where $\left\langle, \right\rangle$ denotes the inner product and $sg( \cdot )$ means stop-gradient. We symmetrize the loss $\mathcal{L}_{\theta, \xi}$ in Eq. \eqref{eq2} by feeding $x'$ to the online network and $x$ to the target network to compute $\tilde{\mathcal{L}}_{\theta, \xi}$. At each training step, a stochastic optimization step is performed to minimize $\mathcal{L}_{BYOL} = \mathcal{L}_{\theta, \xi} + \tilde{\mathcal{L}}_{\theta, \xi}$.

 \begin{algorithm}[t]
  \renewcommand{\algorithmicrequire}{\textbf{Input:}}
  \renewcommand{\algorithmicensure}{\textbf{Output:}}
  \caption{\quad Pseudocode of the Encoder and CMAL}
  \label{alg:code}
  \begin{algorithmic}[0]
     \Require
         $\mathcal{D}$ is a training set
     \Ensure
         $\mathcal{L}_{total}$ is the loss value
 \end{algorithmic}
  \begin{algorithmic}[1]
     \State $\mathcal{L}_{total} \gets 0$
     \For{a minibatch $s$ in $\mathcal{D}$ with $n$ samples}
         \State get the augmented sample $x$ and $x'$
         \State use the online network to obtain the online feature $z_\theta$ and $z_\theta'$ $\gets x, x'$
         \State use the online predictor to obtain the predicted feature $p_\theta$ and $p_\theta'$ $\gets z_\theta, z_\theta'$
         \State use the target network to obtain the online feature $z_\xi$ and $z_\xi'$ $\gets x, x'$
         \State use the mean square error to obtain the loss value in a mini-batch:
         \Statex \mbox{$\mathcal{L}_{CMAL} = mse(p_\theta, p_\theta') - \frac{1}{2}(mse(p_\theta, z_\xi) + mse(p_\theta', z_\xi'))$}
         \State loss backward
         \State update the online network with gradient
         \State update the target network with momentum update
         \State $\mathcal{L}_{total}\gets \mathcal{L}_{total} + \mathcal{L}_{CMAL}$
     \EndFor
     \State \Return $\mathcal{L}_{total}$
  \end{algorithmic}
 \end{algorithm}

\subsection{CMAL}

SkeletonBYOL predicts the output of the target network from an online network. It discards the negative samples completely and achieves good performance. However, it is difficult to understand intuitively why this direct prediction is valid. It is difficult to obtain a more discriminative feature space. Inspired by RAFT \cite{RAFT}, we hope to better use the framework and optimize two goals at the same time: (i) keeping representations of two augmented samples of the same action sequence aligned and (ii) keeping a sample's online representation differing from its target representation. Thus, CMAL is proposed to learn single-stream representation more effectively.

For an input sequence $s$, features $z_\theta$, $z'_\theta$, $z_\xi$ and $z'_\xi$ are extracted. Then prediction head is used to obtain $q_\theta(z_\theta)$ and $q_\theta(z'_\theta)$. In order to obtain a more discriminative feature space and avoid collapse solutions, we first hope to align representations of $x$ and $x'$ similar by ${\mathcal L}_{min}^1$:
\begin{equation}
{\mathcal{L}_{min}^1 = \left\| {{q_\theta }({z_\theta }) - {q_\theta }({{z'}_\theta })} \right\|_2^2}.
\label{eq3}
\end{equation}

On the other hand, we also hope the feature of the online encoder could be different from the target encoder by ${\mathcal L}_{min}^2$ to obtain a more discriminative feature space:
\begin{equation}
{\begin{split}
\mathcal{L}_{min}^2 = \frac{1}{2} &\left(\left\| {{q_\theta }({z_\theta }) - sg({z_\xi })} \right\|_2^2 \right. \\
& + \left. \left\| {{q_\theta }({{z'}_\theta }) - sg({{z'}_\xi })} \right\|_2^2 \right).
\end{split}}
\label{eq4}
\end{equation}

To joint optimize these two goals, our final cross-model adversarial loss is as follows: $\mathcal{L}_{CMAL} = \alpha {\mathcal{L}_{min}^1} + \beta {\mathcal{L}_{min}^2}$, where $\alpha$ and $\beta$ are the coefficients to balance the loss. The two models are updated through momentum, but at the same time, the extracted features are expected to maximize the similarity as much as possible, which forms an implicit confrontation, effectively avoiding model collapse and constructing a more discriminative feature space. The pseudocode of using both the Encoder and the CMAL is in Algorithm \ref{alg:code}.

\subsection{CSCL}

Considering easily-obtained multi-stream skeleton data (i.e., joint, motion, and bone), fusing complementary information preserved in different streams can definitely learn better spatiotemporal information. We design CSCL to mine more ideal feature space in multiple streams.

\textit{(i) Intra-stream similarity.} In single-stream, we gather ${q_\theta }({z_\theta })$, ${q_\theta }({z'_\theta })$, $sg({z_\xi })$, $sg({z'_\xi })$ in a batch and do L2-normalization to get ${Q,Q',K,K'}\in\mathbb{R}^{N\times C}$, respectively. $N$ represents the batch size and $C$ represents the feature dimension. As shown in Fig. \ref{module} (a), we first calculate the cosine similarity in a mini-batch for a single stream. Take ``joint stream'' as an example:
\begin{equation}
{\mathcal{S}_{joint}} = QK'^{\top}, {\mathcal{S'}_{joint}} =  Q'K^{\top},
\label{eq6}
\end{equation}
where ${\mathcal{S}_{joint}}, {\mathcal{S'}_{joint}}\in\mathbb{R}^{N\times N}$. Correspondingly, the similarity matrix $\mathcal{S}_{bone}$, $\mathcal{S}'_{bone}$ of bone stream and $\mathcal{S}_{motion}$, $\mathcal{S}'_{motion}$ of motion stream can also be obtained in a similar way.

\textit{(ii) Intra-stream fusion.} We hope to get a sharper similarity matrix to obtain a more discriminative feature space. As shown in Fig. \ref{module} (b), we use $sharpen( \cdot )$ to make the similarity matrix sharp in a single stream:
\begin{equation}
{\hat {\mathcal{S}}_J} = sharpen({\mathcal S}_{joint}) \; {\rm AND} \; sharpen({\mathcal S}'_{joint}),
\label{eq8}
\end{equation}
here, $sharpen( \cdot )$ operation means setting the diagonal elements and the top-$k$ elements of each row of the similarity matrix to $1$, and the others to $0$ to obtain a sharper similarity matrix. Then perform an ``AND'' operation to obtain the similarity pseudo-label within the stream. In a similar way, we can get ${\hat {\mathcal{S}}_B}$ and ${\hat {\mathcal{S}}_M}$, respectively.

\textit{(iii) Inter-stream voting.} In Fig. \ref{module} (c), we propose to fuse the similarity matrix of three streams to obtain a more ideal similarity matrix ${\hat {\mathcal{S}}} = vote({\hat {\mathcal{S}}_J}, {\hat {\mathcal{S}}_B}, {\hat {\mathcal{S}}_M})$, where $vote( \cdot )$ means ensembling the results to determine the value of each element is $1$ or $0$. Specifically, when there are more than two streams that think the element should be 1, it will be 1, otherwise, it will be 0. Undoubtedly, the result of ensemble learning is better than that of a single stream, so the sharper similarity matrix ${\hat {\mathcal{S}}}$ is more discriminative than ${\hat {\mathcal{S}}_J}$, ${\hat {\mathcal{S}}_B}$, and ${\hat {\mathcal{S}}_M}$.

\textit{(iv) Optimization goal.} A very intuitive idea is to use ${\hat {\mathcal{S}}}$ as a supervision signal to optimize the similarity matrix of each stream, then the optimized similarity matrix of each stream can better generate ${\hat {\mathcal{S}}}$. In this way, information between different streams can be effectively aggregated, thereby obtaining better feature representations. Thus, the similarity difference minimization loss can be calculated like this:
\begin{equation}
{{\cal L}_{CSCL}} =  - \frac{1}{2}\sum\limits {\left[ {\hat {\mathcal{S}}\log ({{\mathcal{S}}_i}) + (1 - \hat {\mathcal{S}})\log (1 - {{\mathcal{S}}_i})} \right]},
\label{eq10}
\end{equation}
where ${\mathcal{S}}_i \in \{{\mathcal{S}_{joint}}, {\mathcal{S'}_{joint}}, {\mathcal{S}_{motion}}, {\mathcal{S'}_{motion}}, {\mathcal{S}_{bone}}, {\mathcal{S'}_{bone}}\}$. This loss is expected to make the feature similarity of every single stream approach the feature similarity of the ensemble feature. Minimizing the difference in similarity rather than minimizing the difference in features is a softer way to obtain a better feature space.

\textbf{Training.} In the early training stage, the encoder of each stream is not stable and strong enough to perform cross-stream information aggregation. Thus, for the CMCS framework, the Encoder is trained with the Cross-Model Adversarial Loss: ${\cal L}_{1} = {\cal L}_{CMAL}$ in the first training stage. Then in the second training stage, the loss function is ${\cal L}_{2} = \lambda {{\cal L}_{CMAL}} + \gamma {{\cal L}_{CSCL}}$ to start aggregating cross-stream information. Here, $\lambda$ and $\gamma$ are the coefficients to balance the loss.

\section{Experiments} \label{section4}

\subsection{Datasets}

\textbf{PKU-MMD Dataset} \cite{pkummd} is composed of two subsets with nearly 20,000 action sequences across 51 action classes. Part \uppercase\expandafter{\romannumeral1} is considered easier for action recognition, while Part \uppercase\expandafter{\romannumeral2} poses more challenges due to significant view variation leading to more skeleton noise. The experiments on these subsets are conducted under the cross-subject protocol.

\textbf{NTU-60 Dataset} \cite{ntu60} includes 56,578 action sequences covering 60 action classes. It has two evaluation protocols: cross-subject (xsub) and cross-view (xview). In xsub, half of the subjects are used for training, and the remaining half are used for testing. In xview, camera 2 and 3 sequences are used for training, while camera 1 sequences are used for testing.

\textbf{NTU-120 Dataset} \cite{ntu120} consists of 120 action classes and 113,945 sequences. It also has two evaluation protocols: cross-subject (xsub) and cross-setup (xset). In xsub, actions performed by 53 subjects are used for training, while the remaining subjects are used for testing. In xset, all 32 setups are split into two halves for training and testing, respectively.

\subsection{Experimental Settings}

\begin{table}[t]
\centering
\caption{Comparison under KNN evaluation protocol with joint stream.}
\label{knn}
\resizebox{\linewidth}{!}{
\begin{tabular}{l|cc|cc|cc}
\toprule
\multirow{2}{*}{Method} & \multicolumn{2}{c|}{PKU-MMD(\%)} & \multicolumn{2}{c|}{NTU-60(\%)} &  \multicolumn{2}{c}{NTU-120(\%)}\\
			                 & part I        & part II       & xsub          & xview         & xsub          & xset \\ \midrule
SkeletonCLR \cite{crossclr}  & 65.9          & 17.9          & 57.7          & 67.3          & 44.9          & 45.8  \\
\textbf{CMAL (Ours)}  & \textbf{77.1} & \textbf{36.6} & \textbf{64.2} & \textbf{72.3} & \textbf{50.0} & \textbf{52.1} \\ \bottomrule
\end{tabular}}
\end{table}

\begin{table*}[t]
\centering
\caption{Linear evaluation results on three datasets.}
\tabcolsep4.5mm
\begin{threeparttable}
\begin{tabular}{c|c|c|cc|cc|c}
\toprule
\multirow{2}{*}{Method} & \multirow{2}{*}{Encoder} & \multirow{2}{*}{Classifier} & \multicolumn{2}{c|}{NTU-120(\%)} & \multicolumn{2}{c|}{PKU-MMD(\%)} & \multirow{2}{*}{Publication\&Year}   \\
& &      & xsub & xset  & part I & part II &                         \\ \midrule
LongTGAN \cite{LongGAN} & GRU                      & FC                                 & -            & -            & 67.7$^\dag$        & 26.0$^\dag$            & AAAI'2018               \\
MS$^2$L \cite{MS2L}        & GRU                      & GRU                                  & -            & -            & 64.9        & 27.6          & ACMMM'2020              \\
P\&C \cite{PandC}      & GRU                      & FC                                & 41.7         & 42.7         & -           & -             & CVPR'2020               \\
PCRP \cite{PCRP}      & GRU                      & FC                                 & 41.7         & 45.1         & -           & -             & TMM'2021               \\
AS-CAL \cite{AS-CAL}    & LSTM                     & FC                                   & 48.6         & 49.2         & -           & -             & Infomation Science'2021 \\
\textbf{CMCS (Ours)} & GRU                  & FC              & \textbf{54.6}         & \textbf{55.0}         & \textbf{84.0}        & \textbf{39.9}            & - \\ \toprule
4s-MG-AL \cite{yang2022motion}         & ST-GCN              & FC            & 46.2            & 49.5            & -           & -             & TCSVT'2022               \\
Cheng \textit{et al.} \cite{HTR}         & Transformer              & FC                                & -            & -            & -           & -             & ICME'2021               \\
MCAE \cite{MCAE}        & MCAE                     & FC                             & 52.8         & 54.7         & -           & -             & NIPS'2021               \\
CP-STN \cite{CP-STN}    & ST-GCN                   & FC                                & 55.7         & 54.7         & -           & -             & ACML'2021               \\
ST-CL \cite{ST-CL}     & GCN                      & FC                               & 54.2         & 55.6         & -           & -             & TMM'2022                \\
SDS-CL \cite{xu2023spatiotemporal}  & DSTA                    & FC                                 & 50.6            & 55.6            & -           & -             & TNNLS'2023               \\
3s-CrosSCLR \cite{crossclr} & ST-GCN                   & FC                                 & 67.9         & 66.7         & 84.9$^\dag$       & 21.2$^\dag$         & CVPR'2021               \\
Yang \textit{et al.} \cite{cloud}  & DGCNN                    & FC                                 & -            & -            & -           & -             & ICCV'2021               \\
Thoker \textit{et al.} \cite{isc} & GRU+CNN                  & FC                                & 67.1         & 67.9         & 80.9        & 36.0            & ACMMM'2020 \\
3s-AimCLR \cite{guo2022contrastive} & ST-GCN                  & FC                                  & 68.2         & 68.8         & 87.8        & 38.5            & AAAI'2022 \\
3s-RVTCLR+ \cite{zhu2023modeling} & ST-GCN & FC  & 68.0 & 68.9 & - & - & CVPR'2023 \\
\textbf{CMCS (Ours)} & ST-GCN    & FC          & \textbf{68.5}     & \textbf{71.1}      & \textbf{88.1}     & \textbf{53.4}            & - \\ \bottomrule
\end{tabular}
\end{threeparttable}
\begin{tablenotes}
\footnotesize
\item{$^\dag$} denotes the result is reproduced according to the provided code.
\end{tablenotes}
\label{com1}
\vspace{-2em}
\end{table*}

\begin{table}[t]
\centering
\caption{Semi-supervised evaluation results.}
\label{semi}
\begin{tabular}{l|cc|cc}
\toprule
\multirow{2}{*}{Method}        & \multicolumn{2}{c|}{PKU-MMD(\%)}     & \multicolumn{2}{c}{NTU-60(\%)} \\
& part \uppercase\expandafter{\romannumeral1} & part \uppercase\expandafter{\romannumeral2}  & xsub  & xview  \\ \midrule
\emph{1\% labeled data:}         &               &                &               &                \\
LongT GAN \cite{LongGAN}         & 35.8          & 12.4           & 35.2          & -              \\
MS$^2$L \cite{MS2L}              & 36.4          & 13.0           & 33.1          & -              \\
ISC \cite{isc}                   & 37.7          & -              & 35.7          & 38.1            \\
3s-Colorization \cite{cloud}     & -             & -              & 48.3          & \textbf{52.5}   \\
3s-CrosSCLR \cite{crossclr}      & 49.7          & 10.2           & 51.1          & 50.0            \\
\textbf{CMCS (Ours)} & \textbf{58.7} & \textbf{17.7}  & \textbf{51.1} & 49.7            \\ \midrule
\emph{10\% labeled data:}        &               &                &               &                 \\
LongT GAN \cite{LongGAN}         & 69.5          & 25.7           & 62.0          & -               \\
MS$^2$L \cite{MS2L}              & 70.3          & 26.1           & 65.2          & -               \\
ISC \cite{isc}                   & 72.1          & -              & 65.9          & 72.5            \\
3s-Colorization \cite{cloud}     & -             & -              & 71.7          & 78.9            \\
3s-CrosSCLR \cite{crossclr}      & 82.9          & 28.6           & 74.4          & 77.8            \\
\textbf{CMCS (Ours)} & \textbf{86.7} & \textbf{37.2}  & \textbf{76.2} & \textbf{79.6}   \\ \bottomrule
\end{tabular}
\vspace{-1em}
\end{table}

\begin{table}[t]
\centering
\caption{Finetuned results on NTU-60 and NTU-120 dataset.}
\label{finetune}
\begin{tabular}{l|cc|cc}
\toprule
\multirow{2}{*}{Method} & \multicolumn{2}{c|}{NTU-60(\%)} & \multicolumn{2}{c}{NTU-120(\%)} \\
                                    & xsub           & xview           & xsub            & xset            \\ \midrule
ST-GCN \cite{ST-GCN}                & 82.8           & 91.0            & 76.4            & 76.8            \\
\textbf{CMAL (Ours)}       & \textbf{83.9}  & \textbf{91.6}   & \textbf{77.4}   & \textbf{79.4}   \\ \midrule
3s-ST-GCN \cite{ST-GCN}             & 85.1           & 91.8            & 80.0            & 80.4            \\
3s-CrosSCLR \cite{crossclr}         & 86.2           & 92.5            & 80.5            & 80.4            \\
3s-AimCLR \cite{guo2022contrastive} & 86.9           & 92.8            & 80.1            & 80.9            \\
\textbf{CMCS (Ours)}    & \textbf{86.9}  & \textbf{92.8}   & \textbf{81.7}   & \textbf{82.7}   \\ \bottomrule
\end{tabular}
\vspace{-1em}
\end{table}

\begin{table}[t]
\centering
\caption{Exploration of different pretext datasets for fine-tuning on PKU-MMD dataset.}
\tabcolsep5mm
\begin{tabular}{c|cc}
\toprule
\multirow{2}{*}{Pretext Task Dataset} & \multicolumn{2}{c}{PKU-MMD(\%)} \\
                    & part I                            & part II          \\ \midrule
w/o pretext task    & 92.3                              & 53.7            \\
PKU-MMD             & 94.1 (\textcolor{blue}{+1.8})     & 58.5 (\textcolor{blue}{+4.8})            \\
NTU-60 xsub         & 95.1 (\textcolor{blue}{+2.8})     & 61.2 (\textcolor{blue}{+7.5})            \\
NTU-120 xsub        & 94.7 (\textcolor{blue}{+2.4})     & 62.4 (\textcolor{blue}{+8.7})            \\ \bottomrule
\end{tabular}
\label{transfer}
\vspace{-1em}
\end{table}

\begin{table}[t]
\centering
\caption{Ablation study results on NTU-60 dataset.}
\label{ab}
\begin{tabular}{l|p{1cm}<{\centering} p{1cm}<{\centering}}
\toprule
\multirow{2}{*}{Method}                             &   \multicolumn{2}{c}{NTU-60(\%)} \\
                                                    & xsub           & xview             \\ \midrule
SkeletonCLR \cite{crossclr}                      & 75.0           & 79.8              \\
SkeletonBYOL (Baseline: Encoder + BYOL)                          & 77.3           & 82.6              \\
CMAL (Encoder + CMAL)       & 78.0           & 83.6              \\
CSCL (Encoder + CSCL)              & 78.5           & 84.2              \\
CMCS (Encoder + CMAL + CSCL) & \textbf{78.6}  & \textbf{84.5}     \\ \bottomrule
\end{tabular}
\vspace{-1em}
\end{table}

\begin{table*}\small
	\centering
\caption{Linear evaluation results compared with SkeletonCLR on NTU-60, PKU-MMD, and NTU-120 datasets.}
	\label{single}
\begin{threeparttable}
	\resizebox{0.8\linewidth}{!}{
	\begin{tabular}{l|c|cc|cc|cc|cc|cc|cc}
		\toprule
		\multirow{3}{*}{Method} & \multirow{3}{*}{Stream} & \multicolumn{4}{c|}{NTU-60(\%)} & \multicolumn{4}{c|}{PKU-MMD(\%)} & \multicolumn{4}{c}{NTU-120(\%)}  \\ \cline{3-14}
		&      & \multicolumn{2}{c|}{xsub}          & \multicolumn{2}{c|}{xview}          & \multicolumn{2}{c|}{part I}       & \multicolumn{2}{c|}{part II}     & \multicolumn{2}{c|}{xsub} & \multicolumn{2}{c}{xset} \\
		&      & acc.            & $\Delta$         & acc.             & $\Delta$         & acc.            & $\Delta$      &acc.&$\Delta$& acc.            & $\Delta$       & acc.              & $\Delta$       \\ \midrule
		SkeletonCLR  &\textbf{J}  &68.3&&76.4&&80.9&&35.2&&56.8&&55.9& \\
		\textbf{CMAL (Ours)} &\textbf{J}& \textbf{73.0} & $\uparrow$ 4.7& \textbf{79.4}& $\uparrow$ 3.0& \textbf{86.3}& $\uparrow$ 5.4 & \textbf{47.7} &$\uparrow$ 12.5 & \textbf{61.6}   & $\uparrow$ 4.8 &   \textbf{63.2}   & $\uparrow$ 7.3 \\ \midrule
		SkeletonCLR   & \textbf{M}   & 53.3  &   & 50.8  &  & 63.4  &  &13.5 &  & 39.6 &  & 40.2 & \\
		\textbf{CMAL (Ours)}     & \textbf{M}&\textbf{57.7}& $\uparrow$ 4.4& \textbf{62.4} & $\uparrow$ 11.6& \textbf{72.1}&$\uparrow$ 8.7 & \textbf{29.1} &$\uparrow$ 15.6 & \textbf{43.5}& $\uparrow$ 3.9 &\textbf{46.2}   & $\uparrow$ 6.0 \\ \midrule
		SkeletonCLR    & \textbf{B}     & 69.4  &  & 67.4  &  & 72.6  &  & 30.4 &  & 48.4 &  & 52.0 & \\
		\textbf{CMAL (Ours)}     & \textbf{B}& \textbf{70.0}& $\uparrow$ 0.6& \textbf{74.0}& $\uparrow$ 6.6& \textbf{83.3} & $\uparrow$ 10.7 & \textbf{34.1} &$\uparrow$ 3.7 & \textbf{59.2}& $\uparrow$ 10.8 &\textbf{59.2}& $\uparrow$ 7.2 \\ \midrule
		3s-SkeletonCLR & \textbf{J}+\textbf{M}+\textbf{B}                 &75.0&&79.8&&85.3&&40.4&&60.7&&62.6& \\
		\textbf{CMAL (Ours)}  & \textbf{J}+\textbf{M}+\textbf{B} &\textbf{78.0}&$\uparrow$ 3.0&\textbf{83.6}&$\uparrow$ 3.8 & \textbf{87.4}&$\uparrow$ 2.1&\textbf{52.8}&$\uparrow$ 12.4&\textbf{66.8}&$\uparrow$ 6.1&\textbf{70.0}&$\uparrow$ 7.4\\ \midrule
3s-CrosSCLR & \textbf{J}+\textbf{M}+\textbf{B}                 &77.8&&83.4&&84.9&&21.2&&67.9&&66.7& \\
		\textbf{CMCS (Ours)}  & \textbf{J}+\textbf{M}+\textbf{B} &\textbf{78.6}&$\uparrow$ 0.8&\textbf{84.5}&$\uparrow$ 0.9 & \textbf{88.1}&$\uparrow$ 3.2&\textbf{53.4}&$\uparrow$ 32.2&\textbf{68.5}& $\uparrow$ 0.6 &\textbf{71.1}&$\uparrow$ 4.4\\ \bottomrule
	\end{tabular}}
\end{threeparttable}
\begin{tablenotes}
\footnotesize
\item{$\Delta$} represents the gain compared to SkeletonCLR using the same stream data. \textbf{J}, \textbf{M} and \textbf{B} indicate joint stream, motion stream, bone stream, respectively. 3s means using three stream.
\end{tablenotes}
\vspace{-1em}
\end{table*}

\textbf{Pretext Training.} All experiments are conducted on the PyTorch framework \cite{pytorch} with a single GeForce GTX 3090 GPU. For data pre-processing, mini-batch size, and encoder, we all follow CrosSCLR \cite{crossclr} for fair comparisons. The representation $y_\theta$ has a feature dimension of 512. The projection head is composed of the following layers: Linear layer with an output size of 2048, Batch normalization layer, Rectified Linear Unit (ReLU) activation function, Final linear layer with an output dimension of 512. The architecture of the prediction head is the same as the projection head. The target decay rate $\tau$ is set to 0.996. For optimization, we use SGD with momentum (0.9), weight decay (0.0001), and a learning rate of 0.1. Since the size of the three datasets varies greatly, it is favourable to set different training epochs for them to avoid overfitting and achieve better convergence. For NTU-60, we set the training epoch to 100 (the first 60 epoch is the first training stage). For NTU-120, we set the training epoch to 80 (the first 40 epoch is the first training stage). For PKU-MMD Part I, we set the training epoch to 150 (the first 100 epoch is the first training stage). For PKU-MMD Part II, we set the training epoch to 250 (the first 200 epoch is the first training stage).

\textbf{Downstream Testing.} We evaluate the trained encoder under multiple protocols after the pretext training.

\textit{(i) KNN Evaluation Protocol.} It involves using a K-Nearest Neighbor (KNN) classifier on the features learned by the trained encoder. It helps assess the quality and effectiveness of the learned features in representing the data. For all reported KNN results, $K = 20$.

\textit{(ii) Linear Evaluation Protocol.} The encoder is evaluated by training a classification head. The encoder is fixed during this process. The classification head consists of a fully connected layer followed by a softmax layer. The training of the classification head is performed using SGD with an initial learning rate of 3.0 for 100 epochs. The learning rate is reduced to 0.3 at epoch 80. The batch size is set to 128. 

\textit{(iii) Semi-supervised Evaluation Protocol.} After training the encoder, a linear classifier is added to the trained encoder. The entire model is then trained using only a small percentage (1\% or 10\%) of randomly selected labeled data. SGD is employed for training the whole network with an initial learning rate of 0.1, which is reduced by a factor of 10 at epoch 80. The entire network is trained for 150 epochs with a batch size set to 128, except for PKU-MMD part II under the 1\% semi-supervised evaluation protocol, where the batch size is set to 52 due to the limited data.

\textit{(iv) Finetune Evaluation Protocol.} Similar to Semi-supervised Evaluation Protocol, but use all labeled data.

\begin{table}[t]
\centering
\caption{Results of different stream on NTU-60 dataset.}
\label{modal}
\begin{threeparttable}
\begin{tabular}{l|l|cc}
\toprule
Method                            & Stream       & xsub(\%)           & xview(\%) \\ \midrule
CMAL                        & J              & 73.0           & 79.4  \\
CMAL                        & M              & 58.8           & 62.9  \\
CMAL                        & B              & 70.1           & 74.0  \\  \midrule
CMAL $^\ddag$          & J + B          & 74.8           & 80.5  \\
CMAL $^\ddag$          & J + M          & 74.5           & 79.3  \\
CMAL $^\ddag$          & M + B          & 73.3           & 78.1  \\
CMAL $^\ddag$          & J + M + B      & 76.2           & 81.4   \\ \midrule
CMCS                           & J + B          & 77.8           & 83.1  \\
CMCS                           & J + M          & 75.4           & 83.9  \\
CMCS                          & M + B          & 76.4           & 81.7  \\
\textbf{CMCS}                  & J + M + B      & \textbf{78.6}  & \textbf{84.5}   \\ \bottomrule
\end{tabular}
\end{threeparttable}
\begin{tablenotes}
\footnotesize
\item{\qquad} \quad $\ddag$ represents directly ensemble the results.
\end{tablenotes}
\vspace{-1em}
\end{table}

\subsection{Comparison with State-of-the-Art Methods}

\textbf{KNN Evaluation Results.} Indeed, the KNN classifier does not involve learning additional weights, making it a simple yet effective evaluation metric for the learned features. The results presented in Table \ref{knn} provide fair comparisons under the specified value of $K$. It is evident that our single-stream CMAL outperforms SkeletonCLR \cite{crossclr} on all three datasets when using the KNN classifier. These substantial improvements achieved with a simple classifier indicate that the features learned by our CMAL model are more discriminative and capture better representations of the action data.

\textbf{Linear Evaluation Results.} As shown in Table \ref{com1}, we compare the proposed CMCS with other recent methods. For using GRU as the encoder, our CMCS outperforms other methods on the three datasets under the linear evaluation protocol. For using stronger encoders, our CMCS also leads 3s-AimCLR under many protocols. For the performance on NTU-120, our CMCS still outperforms other methods by a large margin. The results demonstrate that our CMCS method is highly competitive, especially on multi-class large-scale datasets. On the PKU-MMD dataset, Part II presents more challenges due to increased skeleton noise caused by view variation. In this scenario, our CMCS outperforms 3s-CrosSCLR, highlighting its capability to handle movement patterns arising from skeleton noise effectively. This effectiveness extends to both large-scale and small-scale datasets, indicating the generalization ability of our method. These findings collectively showcase the strong performance and versatility of our CMCS method in various action recognition scenarios.

\textbf{Semi-Supervised Evaluation Results.} Table \ref{semi} shows that even with a small labeled subset, our CMCS outperforms other methods. Remarkably, when using only 1\% and 10\% labeled data, the performance of our CMCS surpasses ISC \cite{isc}, 3s-CrosSCLR \cite{ST-GCN}, and 3s-Colorization \cite{cloud}. These results demonstrate the effectiveness of our approach in leveraging limited labeled data and obtaining superior performance in the semi-supervised learning setting. The learned spatial-temporal representations from the pretext task significantly contribute to achieving these competitive results under such challenging conditions. It also proves that our method effectively learns intra-inter stream information and obtains a more discriminative feature space.

\textbf{Finetuned Evaluation Results.} We have conducted a comparison of our method with supervised ST-GCN and supervised 3s-ST-GCN. These models have the same structure and parameters as our proposed CMCS method. As shown in Table \ref{finetune}, our CMAL achieves better results than supervised ST-GCN, which means that our framework could learn more information to get a better weight initialization. For the results of three streams, the finetuned results also outperform supervised 3s-ST-GCN and finetuned 3s-CrosSCLR, which indicates the effectiveness of our method.

\textbf{Transfer Learning Results.} As a common practice, transfer learning performs self-supervised pre-training on the large-scale ImageNet \cite{imagenet}, then initialize the network with the learned weights and finally train on the small dataset. We also conduct experiments to do pretext tasks on a larger dataset without labels, then finetune on the small dataset. In Table \ref{transfer}, when training from scratch, the accuracy of part I and part II is 92.3\% and 53.7\%. When we do the pretext task on the PKU-MMD dataset itself, we can gain the improvement of 1.8\% and 4.8\%. When using the NTU-60 and NTU-120 datasets for self-supervised pre-training, there is a substantial improvement in accuracy, particularly in part II of these datasets. This significant boost in performance highlights the transferability of the learned representations. 

\subsection{Ablation Study}

\begin{table}[t]
\centering
\caption{Feature dimension search experiments on joint stream of NTU-60 dataset.}
\label{dim}
\begin{tabular}{ccc|ccc}
        	\toprule
        	$d_y$   & $h_{mlp}$  & $d_z$    & xsub(\%)         & xview(\%)         & Avg.(\%)      \\ \midrule
        	256     & 1024       & 256      & 69.4          & 74.0          & 71.7      \\
            256     & 1024       & 512      & 69.4          & 74.3          & \textbf{71.9}      \\
            256     & 1024       & 1024     & 69.7          & 73.7          & 71.7      \\ \midrule
            512     & 1024       & 256      & 70.8          & 76.9          & 73.9      \\
            512     & 1024       & 512      & 72.7          & 75.4          & \textbf{74.1}      \\
            512     & 1024       & 1024     & 71.0          & 77.0          & 74.0      \\ \midrule
            512     & 512        & 512      & 71.3          & 76.7          & 74.0      \\
            512     & 2048       & 512      & \textbf{73.2} & \textbf{78.2} & \textbf{75.7}      \\ \bottomrule
\end{tabular}
\vspace{-1em}
\end{table}

\begin{table}[t]
    \centering
	\parbox{.5\linewidth}{
		\centering
        \caption{Results of different target decay rate $\tau$. (\%)}
        \label{tbl:m}
		\footnotesize
		\setlength{\tabcolsep}{6pt}
        \begin{tabular}{c|cc|c}
        \toprule
        $\tau$  & xsub          & xview         & Avg.        \\ \midrule
        1       & 54.7          & 58.6          & 56.7        \\
        0.999   & 70.8          & 76.1          & 73.4        \\
        0.996   & 73.0          & \textbf{79.4} & \textbf{76.2}        \\
        0.99    & \textbf{73.2} & 78.2          & 75.7       \\
        0.9     & 68.9          & 74.0          & 71.5       \\
        0       & 55.2          & 54.8          & 55.0       \\ \bottomrule
        \end{tabular}\vspace{-1em}
	    }\hfill
    \parbox{.5\linewidth}{
		\centering
        \caption{Results of different Top-$K$ in CSCL on NTU-60 dataset. (\%)}
        \label{k}
		\footnotesize
		\setlength{\tabcolsep}{6pt}
        \begin{tabular}{c|cc}
        \toprule
        Top-$K$  & xsub          & xview         \\ \midrule
        $0$      & 78.0          & 83.6          \\
        $1$      & \textbf{78.7} & 84.4          \\
        $2$      & 78.6          & \textbf{84.5} \\
        $3$      & 78.3          & 83.9          \\
        $4$      & 77.6          & 83.4          \\
        $5$      & 76.6          & 78.5          \\ \bottomrule
        \end{tabular}\vspace{-1em}
		}\hfill
\end{table}

\begin{table}[t]
    \centering
	\parbox{.5\linewidth}{
		\centering
        \caption{Results of different weights to balance $\mathcal{L}_{min}$ and $\mathcal{L}_{max}$. (\%)}
        \label{weight1}
		\footnotesize
		\setlength{\tabcolsep}{6pt}
        \begin{tabular}{cc|cc}
        \toprule
        $\alpha$        & $\beta$     & xsub          & xview        \\ \midrule
        $1$             & $0$         & 8.0           & 7.9          \\
        $0$             & $1$         & 31.9          & 40.4         \\
        $1$             & $0.1$       & 60.3          & 68.6         \\
        $1$             & $1$         & \textbf{73.0} & \textbf{79.4} \\
        $1$             & $10$        & 62.1          & 60.3          \\ \bottomrule
        \end{tabular}
	    }\hfill
    \parbox{.5\linewidth}{
		\centering
        \caption{Results of different weights. (\%)}
        \label{weight2}
		\footnotesize
		\setlength{\tabcolsep}{6pt}
        \begin{tabular}{cc|cc}
        \toprule
        $\lambda$       & $\gamma$    & xsub          & xview        \\ \midrule
        $1$             & $0$         & 78.0          & 83.5          \\
        $0$             & $1$         & 77.4          & 82.3          \\
        $1$             & $0.1$       & 77.2          & 81.5          \\
        $1$             & $1$         & 64.1          & 53.6          \\
        $1$             & $10$        & \textbf{78.6} & \textbf{84.5} \\ \bottomrule
        \end{tabular}
		}\hfill
\end{table}

\begin{table}[h]
\centering
\caption{Results of CMAL on NTU-60 joint stream with different backbone.}
\begin{tabular}{l|c|cc}
\toprule
\multirow{2}{*}{Backbone} & \multirow{2}{*}{Setting} & \multicolumn{2}{c}{NTU-60(\%)} \\
                          &                 & xsub            & xview   \\ \midrule
\multirow{2}{*}{ST-GCN}        & Supervised   & 82.8            & 91.0            \\
                               & Finetuned & \textbf{83.9}   & \textbf{91.6}         \\ \midrule
\multirow{2}{*}{CTR-GCN}       & Supervised   & 84.0            & 91.2            \\
                               & Finetuned & \textbf{85.5}   & \textbf{91.7} \\ \midrule
\multirow{2}{*}{UNIK}          & Supervised   & 83.2            & 89.8            \\
                               & Finetuned & \textbf{84.4}   & \textbf{91.6} \\ \bottomrule
\end{tabular}
\label{backbone}
\vspace{-1em}
\end{table}

\textbf{Effectiveness of CMAL and CSCL.} We also verify the effectiveness of our proposed individual components. In Table \ref{ab}, SkeletonCLR \cite{crossclr} achieves 75.0\% on xsub and 79.8\% on xview while the proposed SkeletonBYOL achieves 77.3\% on xsub and 82.6\% on xview. Our CMAL outperforms these results by 0.7\% and 1.0\%. For the proposed CSCL, applying it to the Encoder also brings significant gains, which also shows the effectiveness of the strategy. For our complete method CMCS, it achieves 78.6\% and 84.5\%.

\begin{figure}[t]
\centering
\subfigure[SkeletonCLR]{\includegraphics[width=2.8cm]{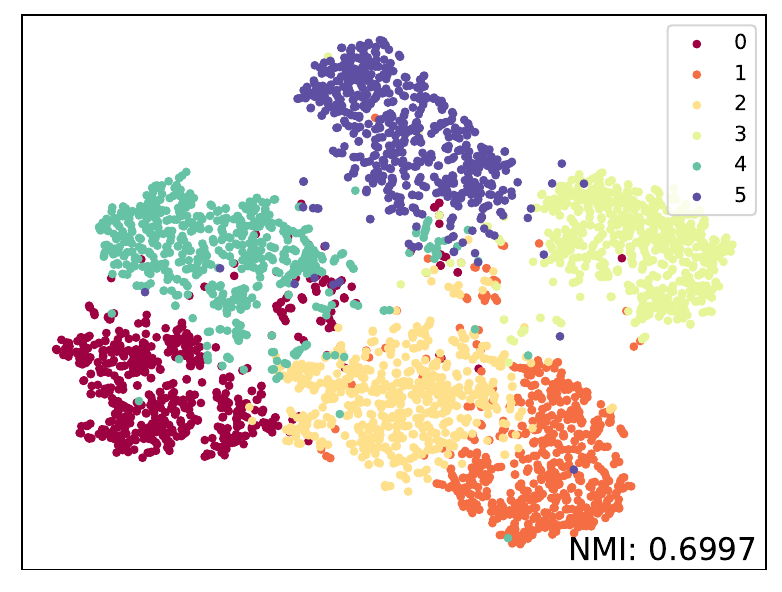}}
\subfigure[SkeletonBYOL]{\includegraphics[width=2.8cm]{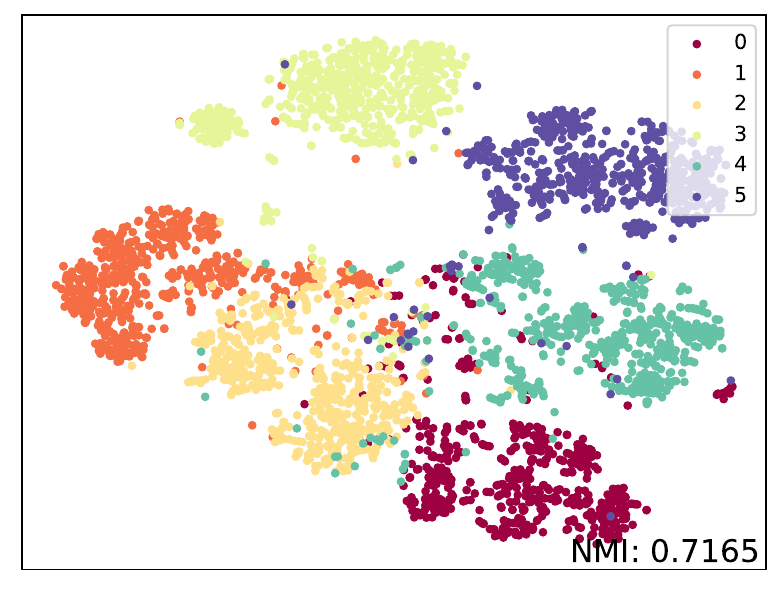}}
\subfigure[CMAL]{\includegraphics[width=2.8cm]{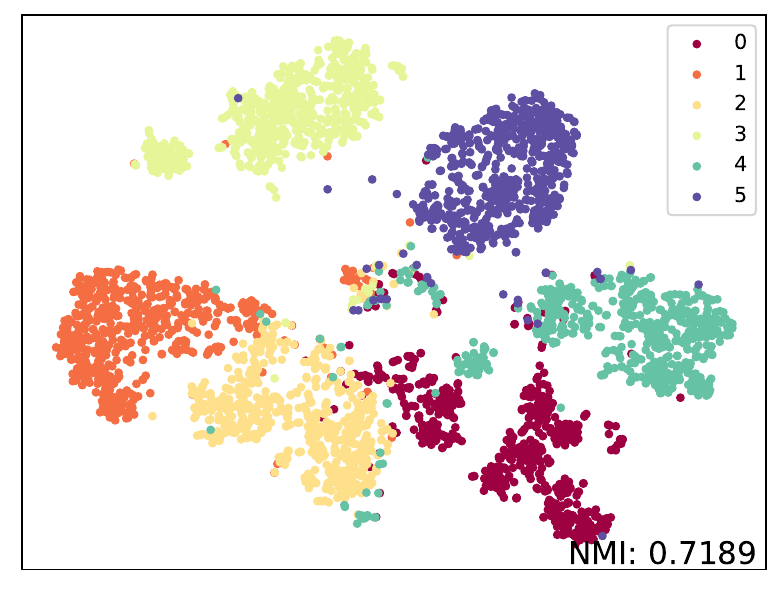}}
\\
\centering
\subfigure[3s-CrosSCLR]{\includegraphics[width=2.8cm]{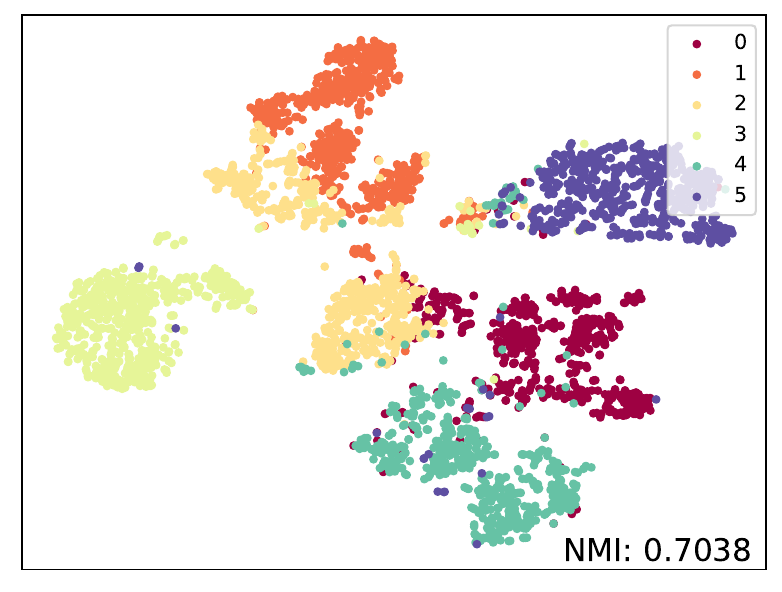}}
\subfigure[3s-CMAL]{\includegraphics[width=2.8cm]{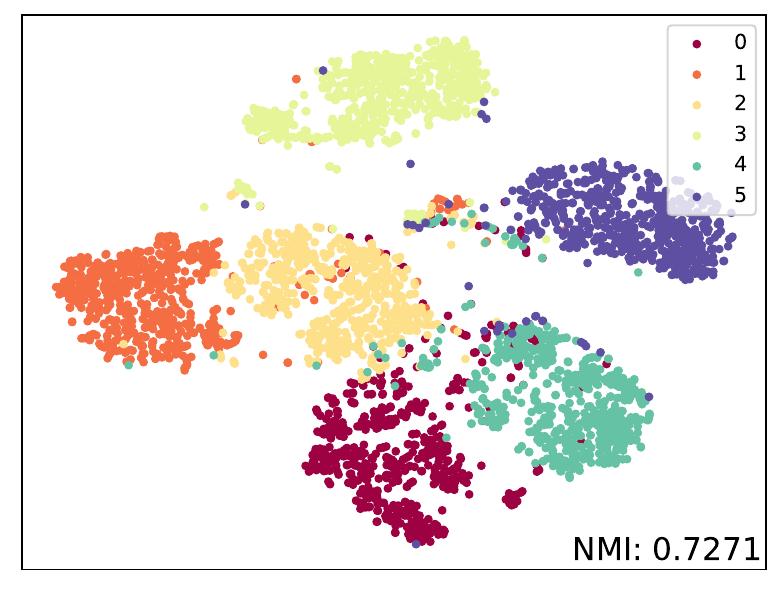}}
\subfigure[CMCS]{\includegraphics[width=2.8cm]{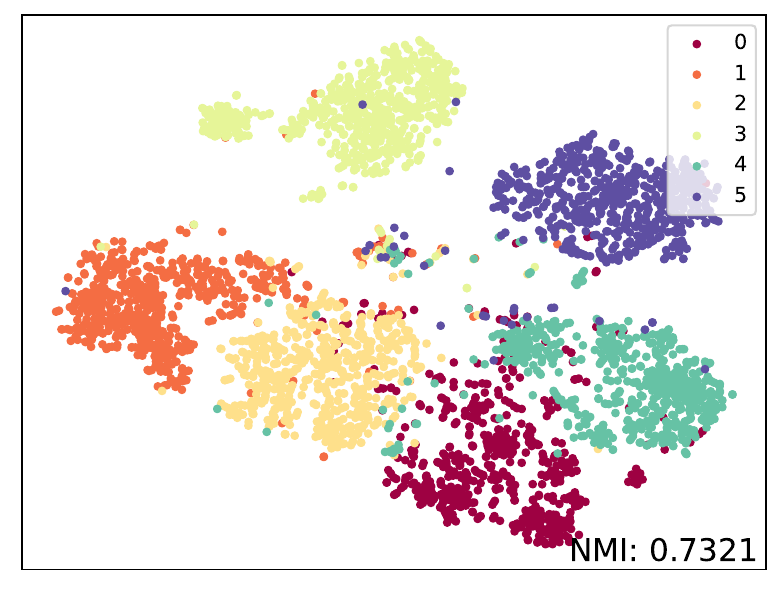}}
\caption{The t-SNE visualization of embeddings on NTU-60 xview. Different methods extract features for 6 categories of samples, and the visualization results after feature dimensionality reduction are shown in the figure. Noting that 3s-CMAL means using CMAL on three streams. {Compared with previous SkeletonCLR and 3s-CrosSCLR methods, our proposed SkeletonBYOL and CMAL do not show obvious improvements, meanwhile our proposed 3s-CMAL and CMCS methods show better performances by separating different colored dots more clearly from each other.}}
\label{tsne}
\end{figure}

\textbf{Comparisons with SkeletonCLR.} We compare our CMAL with the SkeletonCLR in more detail. As shown in Table \ref{single}, for the three different streams (i.e., joint, motion, and bone) of three datasets, our CMAL outperforms SkeletonCLR. For the direct ensemble results, our CMAL also outperforms 3s-SkeletonCLR. It is worth mentioning that both 3s-CrosSCLR and our CMCS explore the information interaction between streams, but our method outperforms 3s-CrosSCLR on three datasets on account that our method better aggregates information from multiple streams. {More ablation on using different streams are explored in Table \ref{modal}}.

\textbf{Choice of Feature Dimension.} Since the predictor has the same structure as the projector, there are three hyper-parameters of feature dimension: the dimension of feature $y$ after encoder $d_y$, the dimension of projector's hidden layer $h_{mlp}$ and the dimension of projector's final layer $d_z$. To determine them, we first fix $h_{mlp}$ and find that the performance of $d_y = 512$ is better than that of $d_y = 256$ as shown in Table \ref{dim}. Then we fix $d_z$ to search for the best $h_{mlp}$. We choose $d_y = 512$, $h_{mlp} = 2048$, $d_z = 512$ as our default settings.

\textbf{Influence of Target Decay Rate $\tau$.} The value of $\tau$ is a trade-off between updating the target network too often and updating too slowly in Table \ref{tbl:m}. When $\tau = 1$, the target network remains constant and is never updated, preserving its initialization value throughout the training process. On the other hand, when $\tau = 0$, the target network is updated instantaneously to match the online network at each step, effectively copying its parameters directly. However, both extreme values, $\tau = 0$ and $\tau = 1$, lead to unstable training and poor performance. To achieve better training stability and performance, the decay rate of $\tau$ is set between 0.9 and 0.999. The value of $\tau = 0.996$ is chosen as it strikes a balance between the stability of training and effective learning, leading to improved performance in the model.

\textbf{Influence of Top-$K$.} From Table \ref{k}, $K = 0$ means that do not use the CSCL. It is worth mentioning that CSCL brings clear gains when $0 < K < 4$. This is because the $Sharpen( \cdot )$ and inter-stream voting retain high confidence samples to learn better feature representations. When $K$ continues to increase, it has a bad impact on performance due to the introduction of too many low-confidence samples. Finally, we take $K$ to 2 according to performance.

\textbf{Influence of Weights in $\mathcal{L}_{CMAL}$.} We also conducted experiments to verify the performance of different weights in $\mathcal{L}_{CMAL}$. In Table \ref{weight1}, using only $\mathcal{L}_{min}$ ($\alpha = 1$, $\beta = 0$) or $\mathcal{L}_{max}$ ($\alpha = 0$, $\beta = 1$) does not perform well. The best performance is achieved when $\mathcal{L}_{min}$ and $\mathcal{L}_{max}$ are equally important ($\alpha = 1$, $\beta = 1$), which is also in line with our original intention when designing cross-model adversarial loss.

\textbf{Influence of Weights to Balance ${\cal L}_{CMAL}$ and ${\cal L}_{CSCL}$.} As shown in Table \ref{weight2}, when do not use the ${\cal L}_{CSCL}$ ($\lambda = 1$, $\gamma = 0$), the performance is also good. However, there is a drop in performance when only using ${\cal L}_{CSCL}$ ($\lambda = 0$, $\gamma = 1$), but not that noticeable. Interestingly, the performance degrades when the weights of ${\cal L}_{CMAL}$ and ${\cal L}_{CSCL}$ are the same. Moreover, for our CMCS, $\lambda = 1$ and $\gamma = 10$ are the best choice. We argue that it is because information aggregation is more important in the second stage of training.

\textbf{Robustness to Different Backbones.} ST-GCN \cite{ST-GCN} is widely used as a backbone for skeleton-based action recognition. To verify the robustness of our method to different backbones, we explore the performance of our method when using a more advanced backbone. We tried CTR-GCN \cite{ctrgcn} (GCN-based) and UNIK \cite{unik} (CNN-based). Due to the limitation of computing resources, for fair comparisons, like ST-GCN, the input frames of CTR-GCN and UNIK are also 50 frames, and the number of channels in the network is also reduced to 1/4 of the original. Due to these changes, the fully supervised results drop somewhat compared to the results in the original paper. But thankfully, as shown in Table \ref{backbone}, under the same experimental setting, our method can improve the performance of end-to-end training. This also illustrates the potential and generalizability of our method to different backbones.

\textbf{Qualitative Results.} In Fig. \ref{tsne}, we visualize the embedding distribution using t-SNE \cite{tsne} with fixed settings. The t-SNE results are reported as fair comparisons, and they are obtained using the same randomly selected 6 class samples from the dataset. From the visual results and the calculated NMI (Normalized Mutual Information), we can draw a conclusion that our SkeletonBYOL and CMAL's embeddings show closer clustering of the same class compared to SkeletonCLR. For multi-streams, even without CA, our 3s-CMAL achieves a more competitive NMI compared to the 3s-CrosSCLR (0.7271 vs 0.7038). Our final CMCS achieves the best NMI.


\section{Conclusion} \label{section5}

This paper focuses on self-supervised representation learning in skeleton-based action recognition and design an CMCS framework to enhance intra-inter stream representations from unlabeled data. The SkeletonBYOL based on the BYOL framework is first constructed. On this basis, CMAL is proposed to leverage cross-model adversarial loss to learn intra-stream representation more effectively. For multi-streams, CSCL is proposed to aggregate inter-stream information to further enhance the representations. Our proposed CMCS framework outperforms state-of-the-art methods across various evaluation protocols, as demonstrated in the experiments.


\bibliographystyle{IEEEtran}
\bibliography{IEEEtran}

\end{document}